%% file: main.tex
\documentclass[10pt,twocolumn,letterpaper]{article}
\usepackage{cvpr}            
\input{preamble}

\definecolor{cvprblue}{rgb}{0.21,0.49,0.74}
\usepackage[pagebackref,breaklinks,colorlinks,citecolor=cvprblue]{hyperref}

\title{NF3DM: Combining Neural Fields and Deformation Models for 3D Non-Rigid Motion Reconstruction}

\author{$^{1}$Aymen Merrouche
\quad
$^{2}$Stefanie Wuhrer
\quad
$^{2}$Edmond Boyer\\
Univ. Grenoble Alpes, CNRS, Inria, Grenoble INP, LJK, France \\
{\tt\small $^{1}$aymerrouche@gmail.com,  $^{2}$name.surname@inria.fr}
}

\begin{document}
\maketitle

\input{./sec/abstract}

\section{Introduction}
\label{sec:introduction}

\input{./sec/introduction}

\section{Related Works}
\label{sec:related_works}

\input{./sec/related_works}

\section{Method}
\label{sec:method}

\input{./sec/method}

\section{Experiments}
\label{sec:experiemnts}
\input{./sec/experiments}

\section{Conclusions}
\label{sec:conclusions}
\input{./sec/conclusions}

\section{Acknowledgements}
We thank David Bojani\'c, Antoine Dumoulin and Rim Rekik for providing us with SMPL fittings for our experiments. We thank Jean-S\'ebastien Franco for helpful discussions.
This work was funded by the ANR project Human4D (ANR-19-CE23-0020).

{
    \small
    \bibliographystyle{ieeenat_fullname}
    \bibliography{main}
}

\input{./sec/X_suppl}

\end{document}

%% file: preamble.tex
\usepackage[dvipsnames]{xcolor}

\usepackage{amsmath}
\usepackage{makecell}
\usepackage{multirow}
\usepackage{float}
\usepackage{mathbbol}
\usepackage{comment}
\usepackage{pifont}
\newcommand{\cmark}{\ding{51}}%
\newcommand{\xmark}{\ding{55}}%
\usepackage{tabularx}
\usepackage{supertabular}
\usepackage{wrapfig}
\usepackage{diagbox}

%% file: sec/abstract.tex
\begin{abstract}

We introduce a novel, data-driven approach for reconstructing temporally coherent 3D motion from unstructured and potentially partial observations of non-rigidly deforming shapes. Our goal is to achieve high-fidelity motion reconstructions for shapes that undergo near-isometric deformations, such as humans wearing loose clothing.
The key novelty of our work lies in its ability to combine implicit shape representations with explicit mesh-based  deformation models, enabling detailed and temporally coherent motion reconstructions without relying on parametric shape models or decoupling shape and motion. Each frame is represented as a neural field decoded from a feature space where observations over time are fused, hence preserving geometric details present in the input data. Temporal coherence is enforced with a near-isometric deformation constraint between adjacent frames that applies to the underlying surface in the neural field. Our method outperforms state-of-the-art approaches, as demonstrated by its application to human and animal motion sequences reconstructed from monocular depth videos.

\end{abstract}  

%% file: sec/introduction.tex
Non-rigid 3D motion reconstruction involves recovering the shape and movement of objects undergoing arbitrary non-rigid motions based on possibly partial visual observations. Given our naturally dynamic world, this task has extensive applications, particularly in digitizing natural scenes for virtual reality and entertainment. Our focus is on reconstructing a moving 3D shape from a monocular depth video.

This problem addresses shape and motion modeling, with existing methods divided based on their approach to modeling these components. Given partial data,~\eg depth maps, two main categories of methodologies have emerged.

\begin{figure}[!ht]
    \centering
    \includegraphics[width=1.0\columnwidth]{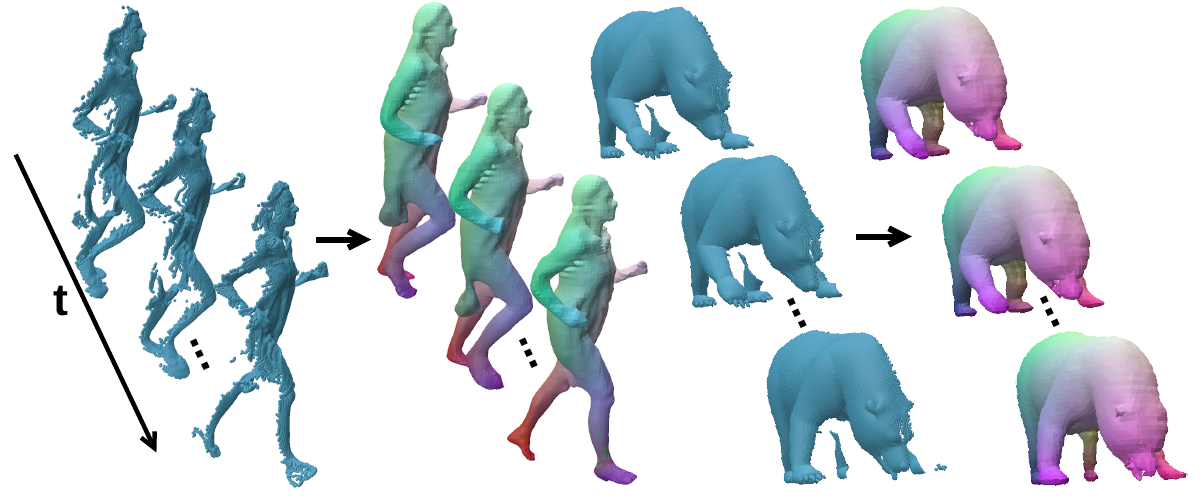}
    \caption{Given monocular depth maps of a moving shape, our approach produces  complete reconstructions  that preserve observed geometric details while establishing dense tracking. We experiment with motions of clothed humans (left) and animals (right).}
    \label{fig:teaser}
\end{figure}

The first one encompasses parametric models, which use combined shape and motion parameters to handle specific entities (\eg humans, faces, or animals). Examples include models like SCAPE~\cite{anguelov2005scape}, BlendSCAPE~\cite{hirshberg2012coregistration}, and SMPL~\cite{loper2023smpl} for human figures, Flame~\cite{li2017learning} for faces
, and SMAL~\cite{zuffi20173d} for animals. These parametric models have gained significant
success, largely thanks to their ability to provide robust, temporally consistent estimations. However, they lack generalizability across shape classes and struggle to capture geometric details outside of model constraints, such as hair or loose clothing in human representations.

The second category includes methods that decouple shape and motion models, allowing for generalization across shape classes. Inspired by Niemeyer~\etal~\cite{niemeyer2019occupancy}, approaches in this category use an implicit scene representation as a template, that undergoes unconstrained displacement fields (e.g., 3D flows) over time. While templates enforce consistency, they can restrict the model's ability to depict finer geometric details. Additionally, 3D flow-based motion modelling strategies are insufficiently constrained, leading to distorted surfaces. Another branch in this category focuses on time-based shape reconstructions rather than motion, either omitting explicit motion modeling altogether~\cite{zhou2021spatio} or limiting it to frame pairs~\cite{li20214dcomplete, zhou2023human}. Although useful for shape capture, these approaches lack broader applicability where motion is required and may yield topologically inconsistent reconstructions.

We propose a novel data-driven representation of 3D motions that combines an implicit neural field shape representation with a near-isometric mesh deformation model. This combination enables tracking a large class of 3D motions such as clothed 3D human motions and other vertebrate motions such as animals. As neural field representations allow for detailed reconstructions and flexible topologies, we leverage the near-isometric deformation to promote temporal coherence and minimal surface distortions, which helps resolve shape inconsistencies that may arise from independent reconstructions.

To this end, we introduce a data-driven approach
that combines two modules working in synergy. A fusion module, that fuses input observations over a time sequence to infer full neural field reconstructions at each time step using an encoder-decoder architecture equipped with an attention mechanism. And a deformation module, that predicts inter-frame deformations by fitting the neural field reconstructions to a near-isometric mesh deformation model using a deformation network. Both modules train together  
without motion supervision but with losses that promote geometric feature association between frames, near-isometric deformations, and 3D reconstruction losses. 
After training, 
per-frame reconstructions and their tracking
are inferred from monocular observations in a single forward step.

To evaluate the approach, we experimented with monocular depth videos of both humans and animals. The results,~\eg Fig.~\ref{fig:teaser}, demonstrate that our method, while exhibiting strong generalization abilities, achieves detailed 3D motion reconstructions that outperform the state of the art.

Our contributions can be summarised as follows:
\begin{itemize}
    \item Investigate and demonstrate the benefits of combining neural field representations with explicit surface deformation models when handling temporal sequences.
    \item A feature-fusion mechanism that generates complete 3D shape reconstructions from potentially partial observations of a 3D shape in motion.
    \item A deformation-guided, unsupervised surface tracking strategy that promotes geometric and topological consistency in the reconstructions.
\end{itemize}

%% file: sec/related_works.tex
Methods to reconstruct a possibly moving 3D shape can be categorized into two classes. On the one hand, array-based methods~\cite{bleyer2011patchmatch, yariv2021volume, wang2021neus, zins2021data, zins2023multi, li2023neuralangelo,guedon2023sugar, Huang2DGS2024} use multiple calibrated cameras that require costly setups and are thus restricted to professional use. On the other hand, depth-fusion-like methods~\cite{innmann2016volumedeform, guo2017real, slavcheva2017killingfusion, yu2017bodyfusion, dou2016fusion4d} enable data obtained from commodity sensors to be used for consumer level applications. An active research direction aims to reconstruct possibly moving 3D shapes from sensor data,~\eg from a single RGB or RGB-D image~\cite{saito2020pifuhd, he2020geo, pesavento2024anim}, from an RGB video~\cite{alldieck2018video, feng2022capturing, alldieck2019learning, li2020monocular, burov2021dynamic}, from depth views~\cite{bozic2021neural,zhou2021spatio,zhou2023human,xue2023nsf} or from point clouds~\cite{niemeyer2019occupancy, tang2021learning}. We review methods that input geometry observations,~\ie depth views or point clouds, as our method considers similar inputs. These approaches follow two main lines of works: model-based methods that leverage parametric shape models, and model-free methods that generalize to multiple shape classes.

\subsection{Model-Based Methods}

Given possibly partial observations of a 3D shape in motion, model-based strategies find the best fit of these observations to the parameter spaces defined by a shape model. Such models have been developed for different shape classes, and we focus here on human body models, which often correspond to shape and pose parameter spaces. 

Early model-based strategies propose optimisation-based techniques. Weiss~\etal~\cite{weiss2011home} fit partial observations to the SCAPE~\cite{anguelov2005scape} human body model. Mosh~\cite{loper2014mosh} uses a sparse set of markers to fit SCAPE~\cite{anguelov2005scape} while Mosh++~\cite{mahmood2019amass} uses SMPL~\cite{loper2023smpl}. To augment the expressivity of parametric human models, several approaches~\cite{pons2017clothcap, alldieck2018detailed, von2018recovering} add vertex displacements on top of SMPL~\cite{loper2023smpl} to model clothes, while DoubleFusion~\cite{yu2018doublefusion} proposes to add a 3D graph. More recently, data-driven approaches were proposed. IP-Net~\cite{bhatnagar2020combining} combines parametric and implicit representations.
H4D~\cite{jiang2022h4d} proposes a compositional representation, which disentangles shape and motion. 
NSF~\cite{xue2023nsf} propose to combine SMPL~\cite{loper2023smpl} with a neural surface field to represent fine grained surface details. 
Neural Parametric Models (NPMs)~\cite{palafox2021npms} propose to learn custom disentangled shape and pose spaces from a dataset to which observations can be fitted  at inference. SPAMs~\cite{palafox2022spams} extend NPMs by learning disentangled semantic-part-based shape and pose spaces.

Unlike these works, our method generalizes to different classes of shapes, including animals and humans with and without clothing. This is achieved using a near-rigid patch-based deformation model to promote geometrically and topologically consistent 3D reconstructions.

\subsection{Model-Free Methods}

\input{./tables/related_works_categories}

We review data-driven model-free methods to reconstruct a possibly moving 3D shape. These methods generalize to different shape classes without needing adjustments, and allow for inference without test-time optimization. For these methods, implicit shape modeling using distances~\cite{park2019deepsdf} or occupancy~\cite{mescheder2019occupancy} became a standard representation. 

Some works consider static 3D reconstruction,~\eg Implicit Feature Networks (IF-Nets)~\cite{chibane2020implicit}. IF-Nets learn to reconstruct an incomplete 3D shape by extracting feature pyramids that retain global and local shape geometry priors.
To allow for dynamic reconstruction, some works complete a sequence of depth observations without computing correspondences over time,~\eg STIF~\cite{zhou2021spatio}. 
4DComplete~\cite{li20214dcomplete} completes the geometry and estimates the motion from one partial geometry and motion field observation. Zhou~\etal~\cite{zhou2023human} complete the geometry and estimate the motion using two time frames containing partial geometric observations of a 3D shape. These works are limited to reconstructing sequences of one or two observations and do not benefit from long-range temporal information.

More related to our work are methods that solve for reconstruction and tracking jointly over long temporal context. Occupancy-Flow (OFlow)~\cite{niemeyer2019occupancy} represents partial observations of a moving 3D shape as an implicit surface undergoing a continuous flow. LPDC~\cite{tang2021learning} uses a spatio-temporal encoder to represent a sequence of point clouds in a latent space that is queried to model the reconstructed frames as occupancy fields with a continuous flow towards the first frame. CaDeX~\cite{lei2022cadex} computes a canonical shape using occupancy that deforms with a homeomorphism to represent a moving 3D shape. Motion2VecSets~\cite{cao2024motion2vecsets} presents a diffusion model to reconstruct 3D motion from noisy or partial point clouds.
These methods factorise a moving 3D shape into a template and 3D flow, which leads to a loss of geometric detail. The lack of constraints on how this flow distorts the moving surface can further alter the deforming surface in occluded areas. 

In contrast, our approach does not decouple shape and motion. Instead, each frame is reconstructed using signed distances, allowing for high fidelity reconstructions. The temporal consistency of these SDFs is constrained using a near-isometric deformation model. As a result, the reconstructions can have high levels of geometric detail while being precisely tracked. Table~\ref{tab:related_works} positions our work~\wrt competing methods according to their ability to provide a dense tracking, complete partial inputs, take into account long temporal context (more than 2 frames), preserve geometric detail present in the input, and train without inter-frame correspondence supervision. Our method is the only one that fulfills all five desiderata.

%% file: tables/related_works_categories.tex
\begin{table}[h!]
\centering
\resizebox{1.0\columnwidth}{!}{
\begin{tabular}{ |c|ccccc|}
 \hline
 Method & \thead{Tracking} & \thead{Shape \\ Completion} & \thead{Long \\ Temp. Ctxt.} & \thead{Detail \\ Preservation } & \thead{Unsupervised} \\

\hline
OFlow \cite{niemeyer2019occupancy}& \cmark &  \cmark & \cmark & \xmark & \cmark \\
LPDC \cite{tang2021learning}& \cmark &  \cmark & \cmark & \xmark & \xmark \\
CaDeX \cite{lei2022cadex}& \cmark &  \cmark & \cmark & \xmark & \cmark \\
Motion2VecSets \cite{cao2024motion2vecsets}& \cmark &  \cmark & \cmark & \xmark & \xmark \\
\hline

Ours  & \cmark & \cmark & \cmark & \cmark & \cmark    \\
\hline
\end{tabular}}
\caption{Classification of related methods~\wrt their ability to provide tracking, handle partial inputs, exploit long temporal context, preserve geometric details of the observations, and train without inter-frame correspondence supervision.}
\label{tab:related_works}
\end{table}

%% file: sec/method.tex
\begin{figure*}[h]
    \centering
    \includegraphics[width=1.0\textwidth]{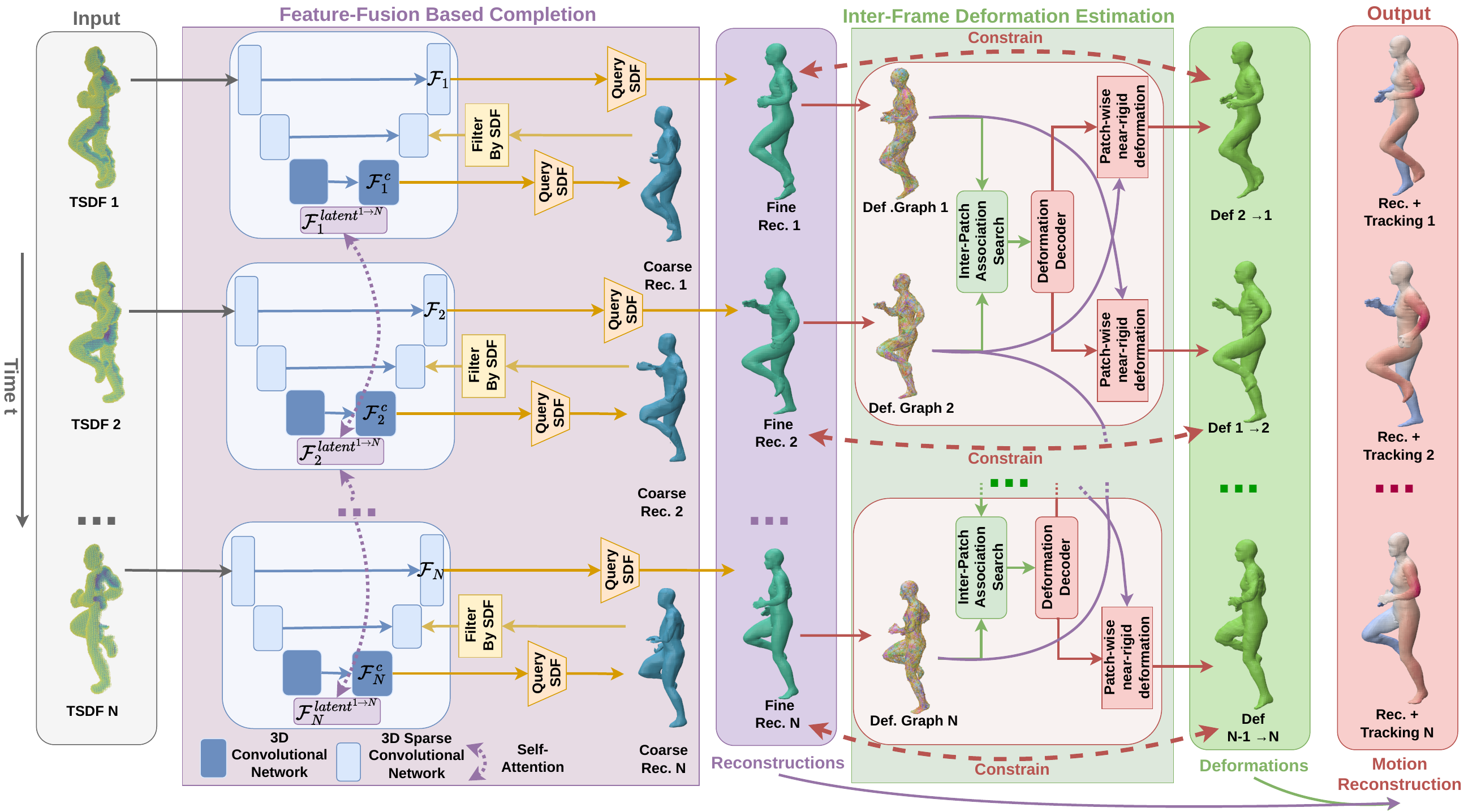}
    \caption{Overview of our approach. Given Truncated Signed Distance Field grids representing partial observations of a moving 3D shape (leftmost), the method achieves detailed reconstructions with dense tracking (rightmost). During Feature-Fusion Based Completion (purple module), the TSDFs are encoded in a latent space where self-attention allows to fuse and complete the observed information. The fused latent features are decoded into coarse shapes and then refined where this coarse surface locates. During Inter-Frame Deformation Estimation (green module), the fusion is further constrained by fitting the  reconstructions to a patch-wise near-rigid mesh deformation model that implements a near-isometric deformation assumption promoting their consistency.}
    \label{fig:method_overview}
\end{figure*}

Given partial observations of a moving 3D shape, our approach estimates both complete shapes and their temporal evolution. To model the latter, recent methods rely on a parametric shape model limiting generalisation to different shape classes, or on an unconstrained 3D flow that deforms a template leading to distortions and to a low preservation of observed geometric detail. Instead, we use a near-isometric mesh deformation model. On the one hand, the near-isometric deformation assumption allows to represent a variety of moving shapes. On the other hand, mesh deformation modeling allows to control the amount of surface distortions induced by the deformation, promoting consistency of the reconstructions. Furthermore, shape and motion are not decoupled. Instead, we propose a multi-scale feature-fusion strategy to represent each frame as a neural field capable of capturing observed geometric details, and link these neural fields under the mesh deformation constraint. Fig.~\ref{fig:method_overview} gives an overview of the method.

Our approach relies on the synergy of two modules. A feature-fusion and completion module (Sec.~\ref{sec:feature_fusion_com}), where the observations are encoded in a latent space, fused and then decoded into complete shapes. To efficiently produce high fidelity completions, we use a multi-scale implicit surface representation using signed distances (purple module in Fig.~\ref{fig:method_overview}). The fusion module trains with complete shape supervision to build a shape space for the completion task.
The fusion is further constrained by learning a surface deformation prior with a deformation search module (Sec.~\ref{sec:inter_frame_def_est}) that translates the fusion and completion in feature space to a near-isometric deformation in 3D. Implicit surfaces are fitted to a mesh-based near-isometric deformation model (green module in Fig.~\ref{fig:method_overview}). Both modules are inter-dependent benefiting one another. With the near-isometric deformation assumption, our method trains without inter-frame correspondences and without requiring a shape in a canonical pose per sequence. It optimises for a fusion objective and for a self-supervised deformation objective: 
\begin{equation}
    \label{total_loss}
    l^{network} = l^{fusion} + l^{def},
\end{equation}
with fusion and deformation objectives $l^{fusion}$ and $l^{def}$.

\subsection{Feature-Fusion Based Completion}
\label{sec:feature_fusion_com}

Given a sequence of $N$ TSDF (truncated signed distance field) volumes $( \mathcal{T}_i \in \mathbb{R}^{D\times H \times W})_{i \in \{1,...,N\}}$, which can be readily obtained from depth videos with known camera parameters~\cite{curless1996volumetric, newcombe2011kinectfusion} and represent partial observations of a moving 3D shape, the feature-fusion based completion fuses these observations in a feature space to complete each frame while retaining observed geometric details. It computes a neural signed distance field $SDF_{\Theta}(i)$ for each frame $i$. To produce high fidelity reconstructions, $SDF_{\Theta}(i)$ is represented with geometry aligned features~\cite{he2020geo, li20214dcomplete, zhou2023human}.
These features are obtained at two scale levels. First  a coarse grid $(\mathcal{F}^c_i \in \mathbb{R}^{D_{c} \times H_{c} \times W_{c} \times C})_{i \in \{1,...,N\}}$ is extracted, that represents coarse completions $SDF^{coarse}_{\Theta}(i)$. Second, with the aim to capture high frequency details,  this grid is refined only where the coarse surface locates, \ie where $SDF^{coarse}_{\Theta}(i)$ is lower than a certain threshold. The refined features are denoted as $(\mathcal{F}_i \in \mathbb{R}^{D_{F} \times H_{F} \times W_{F} \times C})_{i \in \{1,...,N\}}$.

To obtain the SDF value at a query point $x$ for the completion of frame $i$, we first interpolate the feature volume $\mathcal{F}_i$ using trilinear interpolation. The resulting feature, concatenated with $x$, is then passed to a MLP, denoted as $S_{\Sigma}$, to produce the desired value. Given that each $\mathcal{T}_i$ is normalised in a bounding box $B$, we can express $SDF_{\Theta}(i)$ as follows:
\begin{align}
\begin{split}
\scriptstyle
SDF_{\Theta}\colon \{1,...,N\} \times \mathbb{R}^{N \times D\times H \times W}  \times B & \rightarrow \mathbb{R}\\
i, (\mathcal{T}_i)_{i \in \{1,...,N\}}, x &\mapsto  S_{\Sigma}(\operatorname{tri}(\mathcal{F}_i, x),x),
\end{split}
\end{align}
where $\operatorname{tri}$ stands for trilinear interpolation. The same applies to $SDF^{coarse}_{\Theta}(i)$. At training, $l^{fusion}$ in Eq.~\ref{total_loss} leverages complete shape information to supervise the completion at both scales:
\begin{equation}
    \label{fusion_loss}
    l^{fusion} = l^{coarse} + l^{fine},
\end{equation}
with coarse-scale and fine-scale objectives $l^{coarse}$ and $l^{fine}$.

The geometry aligned features are extracted using a feature extractor $F_{\Psi}$ acting in four steps. First, it encodes each $\mathcal{T}_i$ in a latent feature space (Sec.~\ref{sec:latent_encoder}). Then, using an attention mechanism, it fuses the latent features to share the observed information at each frame (Sec.~\ref{sec:fusion}). The fused latent features are decoded into coarse feature volumes representing coarse completions (Sec.~\ref{sec:coarse_rec}). To get high fidelity reconstructions, the coarse feature volumes are refined where the coarse surface locates, using features describing fine details retained during encoding (Sec.~\ref{sec:refinement}).

\subsubsection{Frame-Wise Latent Feature Encoder} 
\label{sec:latent_encoder}
$F_{\Psi}$ first encodes the information observed in each $\mathcal{T}_i$ in a latent feature space. For efficiency, we leverage a sparse convolutional encoder~\cite{3DSemanticSegmentationWithSubmanifoldSparseConvNet} (\ie convolutions are only applied at grid locations where $\mathcal{T}_i$ is defined). This reduces the feature grid to a coarser spatial resolution $D_c\times H_c \times W_c$ where standard 3D convolution is  computationally feasible. A 3D convolutional encoder then processes this grid to produce the latent per-frame features $(\mathcal{F}^{latent}_i \in \mathbb{R}^{d_l})_{i \in \{1,...,N\}}$.

\subsubsection{Feature Fusion}
\label{sec:fusion}
In order to allow for geometric fusion and completion, $F_{\Psi}$ communicates the information encoded in the latent features between frames. This is done thanks to a self-attention mechanism~\cite{vaswani2017attention} applied on the latent codes $\mathcal{F}^{latent}_i$. This means that the latent features outputted by this self-attention mechanism that we denote $(\mathcal{F}^{latent^{1 \rightarrow N}}_i \in \mathbb{R}^{d_l})_{i \in \{1,...,N\}}$ encode fused and completed shape information.

\subsubsection{Coarse-Dense Reconstruction}
\label{sec:coarse_rec}
The latent features encoding fused and completed shapes must be decoded into feature volumes capable of capturing observed geometric details. Inspired by~\cite{li20214dcomplete, zhou2023human}, we use a two-scale grid of features to define each neural field. The underlying surface is first located at a coarse resolution, and then features are refined where this surface locates. The coarse-dense reconstruction step locates this coarse surface. This translates to finding coarse feature volumes $(\mathcal{F}^{c}_i\in \mathbb{R}^{D_c\times H_c \times W_c \times C})_{i \in \{1,...,N\}}$. Since the locations of unobserved shape parts within the bounding box $B$, are unknown, we must obtain dense features, \ie features present throughout $B$, allowing interpolation at any spatial location $x$ to retrieve the associated SDF value. To achieve this, we employ a 3D convolutional decoder on the latent codes  $(\mathcal{F}^{latent^{1 \rightarrow N}}_i)_{i \in \{1,...,N\}}$ to compute the coarse-dense feature volumes $(\mathcal{F}^{c}_i)_{i \in \{1,...,N\}}$. As training objectives, we enforce that the neural fields encoded by these features approximate the ground truth SDF values while preserving the SDF property, using the following losses:

{\footnotesize
\begin{equation}
\label{coarse_loss}
    l^{coarse} =  \lambda_1 l^{SDF}_{coarse} + \lambda_2 l^{eikonal}_{coarse} \mbox{, with} 
\end{equation}
}

{\footnotesize
\begin{equation}
\label{sdf_coarse_loss}
    l^{SDF}_{coarse} =  \frac{1}{N} \sum\limits_{i = 1}^{N} \frac{1}{S} \sum\limits_{j = 1}^{S} (|S_{\Sigma}(\operatorname{tri}(\mathcal{F}^{c}_i , x_j), x_j)   - gt^{i}_{sdf}(x_j) |)
\end{equation}}

{\footnotesize
\begin{equation}
\label{eik_coarse_loss}
    l^{eikonal}_{coarse} =  \frac{1}{N} \sum\limits_{i = 1}^{N} \frac{1}{S} \sum\limits_{j = 1}^{S} (\|\|\nabla_{x_j}S_{\Sigma}(\operatorname{tri}(\mathcal{F}^{c}_i , x_j), x_j)\|_2   - 1\|_2^2)
\end{equation}}

where $\lambda_1,\lambda_2 \in \mathbb{R}$ are weights for loss terms; $(x_j \in B)_{j \in \{1,...,S\}}$ are $S$ points sampled in $B$ and $gt^{i}_{sdf}(x_j)$ is the ground truth SDF value at point $x_j$ for frame $i$ .

\subsubsection{Reconstruction Refinement}
\label{sec:refinement}
We aim to achieve high-fidelity reconstructions that capture fine-grained geometric details. To this end, the coarse-dense feature volumes $(\mathcal{F}^{c}_i)_{i \in \{1,...,N\}}$ are refined at the coarse surface locations using a sparse convolutional decoder~\cite{3DSemanticSegmentationWithSubmanifoldSparseConvNet, li20214dcomplete}. During decoding, fine-grained features describing the observed surface, retained by the encoder, are combined. This enables high-fidelity reconstructions.  We denote these refined features $(\mathcal{F}_i \in \mathbb{R}^{D_F \times H_F \times W_F \times C})_{i \in \{1,...,N\}}$. As training objectives, we use the same losses as defined in Eq.~\ref{sdf_coarse_loss} and in Eq.~\ref{eik_coarse_loss} after replacing $(\mathcal{F}^c_i \in \mathbb{R}^{D_c \times H_c \times W_c \times C})_{i \in \{1,...,N\}}$ with $(\mathcal{F}_i \in \mathbb{R}^{D_F \times H_F \times W_F \times C})_{i \in \{1,...,N\}}$ :

{\footnotesize
\begin{equation}
\label{fine_loss}
    l^{fine} =  \lambda_1 l^{SDF}_{fine} + \lambda_2 l^{eikonal}_{fine},
\end{equation}}

where $\lambda_1, \lambda_2 \in \mathbb{R}$ are weights for loss terms.

\subsection{Inter-Frame Deformation Estimation}
\label{sec:inter_frame_def_est}

As neural fields allow for changes of both geometry and topology during training, we leverage the constraint of a near-isometric mesh deformation model~\cite{cagniart2010free} to promote topological consistency in a data-driven way. We achieve this by checking that the surface underlying the neural field of each frame can be deformed using this model to get the surface underlying its adjacent frames by optimising for this deformation. This translates the fusion and completion made by $F_{\Psi}$ to a near-isometric deformation in 3D.

One challenge is to unify the representations: the completion acts in a volume, while the deformation acts on a surface. To characterise the surface underlying the geometry aligned features $(\mathcal{F}_i)_{i \in \{1,...,N\}}$, we extract the zero-level set of the neural fields using marching cubes~\cite{conf/siggraph/LorensenC87}. This gives a mesh $\mathcal{M}_i$ for each frame $i$. The deformation search between neural distances defined by $\mathcal{F}_i$ and $\mathcal{F}_j$ boils down to a deformation search between meshes $\mathcal{M}_i$ and $\mathcal{M}_j$. Both representations are then linked using training objectives.

The mesh deformation model we consider decomposes a non-rigid deformation into patch-wise rigid deformations, \ie a rotation matrix and a translation vector. The latter are blended at the vertex level to obtain the non-rigid deformation. Each mesh $\mathcal{M}_i$ is thus decomposed into non-overlapping surface patches $(P^i_k)_{1 \leq k \leq L}$ with their centers $C^i = (c^i_k \in \mathbb{R}^3)_{1 \leq k \leq L}$ where $L$ is the number of patches. 

To learn the deformation, we use a patch-wise deformation decoder that inputs associations between patches of $\mathcal{M}_i$ and patches of $\mathcal{M}_j$ and outputs rotation and translation parameters that achieve the input associations~\cite{Merrouche_2023_BMVC}. So, the deformation search acts in two steps: an association estimation step (Sec.~\ref{sec:association_search}) and a deformation estimation step (Sec.~\ref{sec:def_serach}). The inter-frame deformation loss $l^{def}$ in Eq.~\ref{total_loss} is composed of an association term and a deformation term: 

{\begin{equation}
\label{l_def}
    l^{def} =   l^{associate} + l^{deform}.
\end{equation}}

Both $l^{associate}$ and $l^{deform}$ are defined without using inter-frame correspondence supervision and detailed below.

\subsubsection{Inter-Frame Association Search}
\label{sec:association_search}
Given two meshes $\mathcal{M}_i$ and $\mathcal{M}_j$ representing neural fields encoded by $\mathcal{F}_i$ and $\mathcal{F}_j$, we estimate inter-patch associations as association matrices. We first obtain a feature representative of each patch by trilinearly interpolating  $\mathcal{F}_i$ and $\mathcal{F}_j$ at the center of the patches \ie~at $C^i$ and $C^j$ respectively. This gives for meshes $\mathcal{M}_i$ and $\mathcal{M}_j$, a feature for each of their patches denoted $\mathcal{F}^{patch}_i \in \mathbb{R}^{L \times C}$ and $\mathcal{F}^{patch}_j \in \mathbb{R}^{L \times C}$. Following feature similarity based shape matching methods~\cite{eisenberger2021neuromorph, lang2021dpc,Merrouche_2023_BMVC}, we use the cosine similarity of these features to estimate inter-patch association matrices as follows:

\noindent\begin{minipage}{.5\linewidth}
{\footnotesize 
\begin{equation}
\label{pi_x_y}
	{(\Pi_{i \rightarrow j})_{mn}}
	:=
	\frac
	{e^{s_{mn}}}
	{\sum\limits_{k=1}^{L}e^{s_{mk}}}
\end{equation}
}

\end{minipage}%
\begin{minipage}{.5\linewidth}
{\footnotesize 
\begin{equation}
\label{pi_y_x}
	{(\Pi_{j \rightarrow i})_{mn}}
	:=
	\frac
	{e^{s_{nm}}}
	{\sum\limits_{k=1}^{L}e^{s_{km}}}
\end{equation}
}
\end{minipage}

with $s_{mn} := \frac{\langle{\mathcal{F}}^{patch}_{i,m},{\mathcal{F}}^{patch}_{j,n}\rangle_2}{	\|{\mathcal{F}}^{patch}_{i,m}\|_2\|{\mathcal{F}}^{patch}_{j,n}\|_2}$.

For efficiency, we only compute associations and deformations between adjacent frames. We leverage two criteria on the association matrices. First, a cycle consistency criterion~\cite{Merrouche_2023_BMVC, groueix2019unsupervised} that promotes cycle consistent associations, \ie~every point that goes through a cycle is mapped back to itself. We enforce length $2$ and length $3$ cycle consistency for each sequence, ensuring consistency for every cycle~\cite{nguyen2011optimization, groueix2019unsupervised}. Second, we use a self-reconstruction criterion $l^{rec}$ to identify each patch in feature space, in order to avoid many-to-one patch associations~\cite{Merrouche_2023_BMVC}. 

The combination of $l^{cycle}$ and $l^{rec}$ defines $l^{associate}$ in Eq.~\ref{l_def}:

\begin{equation}
\label{l_associoate}
    l^{associate} =   \lambda_3l^{cycle} + \lambda_4l^{rec},
\end{equation}

where $\lambda_3, \lambda_4 \in \mathbb{R}$ are weights for loss terms. Both $l^{cycle}$ and $l^{rec}$ are detailed in Sec.~\ref{sec:assos_losses_supp}.

\subsubsection{Deformation Search}
\label{sec:def_serach}

The association matrices between meshes $\mathcal{M}_i$ and $\mathcal{M}_j$ induce the desired deformation. These matrices deform $\mathcal{M}_i$ (resp. $\mathcal{M}_j$) by mapping the center of its patches from ${C}^i$ (resp. ${C}^j$) to $\Pi_{i \rightarrow j} {C}^j$ (resp. $\Pi_{j \rightarrow i} {C}^i$). 

The associations were obtained from the geometry aligned features $(\mathcal{F}_i)_{i \in \{1,...,N\}}$ without any manipulation, hence, \emph{the geometry aligned features not only define the geometry when queried through $S_{\Sigma}$, but also define the deformations between the reconstructed surfaces}.

 To learn this deformation, we employ a deformation decoder as introduced in~\cite{Merrouche_2023_BMVC} for static, complete 3D shapes. It consists of a graph convolutional network acting on the patch neighborhoods followed by an MLP. It outputs rotation parameters $(R_k \in \mathbb{R}^6)_{1 \leq k \leq L}$ and new center positions $(u_k\in \mathbb{R}^3)_{1 \leq k \leq L}$ for every patch of $\mathcal{M}_i$. Applying this deformation leads to the deformed shape $\mathcal{M}_{i\rightarrow j}$.

The deformation network is trained using three self-supervised criteria. First, the matching loss $l^{match}$ encourages the deformation network to match the association matrices by producing the deformations they induce :

{\footnotesize
\begin{equation}
    \label{matching_loss}
\begin{split}
    l^{match} &= \frac{1}{N-1}(\sum\limits_{i=1}^{N-1}\|C^{i \rightarrow i+1} - \Pi_{i \rightarrow i+1} C^{i+1}\|^2_2 
    \\&+ \sum\limits_{i=2}^{N}\|C^{i \rightarrow i-1} - \Pi_{i \rightarrow i-1} C^{i-1}\|^2_2),
\end{split}
\end{equation}
}

where $C^{i \rightarrow i+1}$ and $C^{i \rightarrow i-1}$  are the deformed patch centers of $\mathcal{M}_{i\rightarrow i+1}$ and $\mathcal{M}_{i\rightarrow i-1}$ respectively. 

Second, the rigidity criterion promotes preserving the continuity of the deformed shapes along the patch borders:  

{\footnotesize
\begin{equation}
    \label{rig_loss}
    l^{rigidity} =\frac{1}{N-1}( \sum\limits_{i=1}^{N-1}l_{rig}(\mathcal{M}_{i\rightarrow i+1}) + \sum\limits_{i=2}^{N}l_{rig}(\mathcal{M}_{i\rightarrow i-1})),
\end{equation}
}

where $l_{rig}$ is the deformation model's rigidity loss detailed in Sec.~\ref{sec:def_model_supp}. The rigidity criterion favors deformations that preserve the intrinsic properties of the mesh \ie~ $l^{rigidity}$ implements the near-isometric assumption.

Third, the surface loss $l^{surf}$ explicitly links the deforming surfaces with the geometry aligned features. It ensures that the deformation network brings the deformed vertices closer to the surface of the target frame, by encouraging the surface of $\mathcal{M}_{i\rightarrow i+1}$ (resp. $\mathcal{M}_{i\rightarrow i-1}$) to lay on the zero level set of the neural field of frame $i+1$ (resp. frame $i-1$):

{\scriptsize
\begin{align}
\label{surface_loss}
    &l^{surf} =  \frac{1}{N-1}(\sum\limits_{i = 1}^{N-1}( \frac{1}{|V(\mathcal{M}_{i \rightarrow i+1})|} \sum\limits_{\substack{v \in \\ V(\mathcal{M}_{i \rightarrow i+1})}}|S_{\Sigma}(\operatorname{tri}(\mathcal{F}_{i+1} , v), v)|) \nonumber\\
    &+ \sum\limits_{i = 2}^{N}(\frac{1}{|V(\mathcal{M}_{i \rightarrow i-1})|} \sum\limits_{\substack{v \in \\ V(\mathcal{M}_{i \rightarrow i-1})}}|S_{\Sigma}(\operatorname{tri}(\mathcal{F}_{i-1} , v), v)|)),
\end{align}}

where $V(\mathcal{M}_{i \rightarrow i+1})$ (resp. $V(\mathcal{M}_{i \rightarrow i-1})$) are the vertices of deformed mesh $\mathcal{M}_{i \rightarrow i+1}$ (resp. $\mathcal{M}_{i \rightarrow i-1}$ ).

Combining the three losses defines $l^{deform}$ in Eq.~\ref{l_def}:

{
\begin{equation}
\label{l_deform}
    l^{deform} =   \lambda_5 l^{match} + \lambda_6 l^{rigidity} + \lambda_7 l^{surf},
\end{equation}
}

where $\lambda_5, \lambda_6, \lambda_7 \in \mathbb{R}$ are weights for loss terms. 

Our network is trained to optimise for $l^{network}$ until convergence. After training, it computes reconstructions and tracking in a single forward pass.

%% file: sec/experiments.tex
We conduct a comparative study on 3D motion reconstruction from monocular depth, aiming to recover complete shapes with dense tracking. We test on clothed and naked humans (Sec.~\ref{sec:human_rec}) and animals (Sec.~\ref{sec:animal_rec}). Ablations assess the method's core components (Sec.~\ref{sec:ablation}). The supplementary presents additional results in Sec.~\ref{sec:npm_comparison} and implementation details in Sec.~\ref{sec:implementation_details}.

\noindent\textbf{Competing Methods} We compare with OFlow~\cite{niemeyer2019occupancy}, LPDC~\cite{tang2021learning}, CaDeX~\cite{lei2022cadex} and Motion2VecSets~\cite{cao2024motion2vecsets}.
Methods able to train without inter-frame correspondence supervision: OFlow, CaDeX, and Ours, are trained unsupervised. 

\noindent\textbf{Evaluation Datasets}
All methods are trained on the Dynamic FAUST (D-FAUST)~\cite{bogo2017dynamic} dataset, consisting of sequences of minimally dressed, aligned and complete human motion sequences. It includes 10 subjects and 129 sequences. We use the train/val/test split introduced in~\cite{niemeyer2019occupancy}. To evaluate cross-dataset generalisation, we evaluate the models trained on D-FAUST on two other test sets. First, on a subset of 4DHumanOutfit~\cite{armando20234dhumanoutfit} which consists of sequences of real clothed human motions captured using a multi-camera platform. Our subset includes 4 subjects and 8 sequences. Second, on a subset of CAPE~\cite{ma2020learning} which consists of sequences of aligned, complete, and clothed human motions. Our subset includes 3 subjects and 12 sequences. 

To evaluate generalisation to other shape classes, all methods are trained on DeformingThings4D-Animals (DT4D-A)~\cite{li20214dcomplete}.  It consists of animations of animal shapes including $38$ animal identities and $1227$ animations. We use the train/val/test split introduced in~\cite{lei2022cadex}.

Similar to~\eg~\cite{lei2022cadex, cao2024motion2vecsets}, we generate synthetic monocular depth videos from the mesh sequences. We use back-projected depth point-clouds ($10k$ points) for the competing methods and compute TSDFs for our approach; hence, all methods input densely sampled surface information. For fair comparisons, all methods are re-trained as we generate depth videos that differ from those used in the original works and train on denser point clouds. Still, our results are consistent with the ones reported in the original works.

\noindent\textbf{Evaluation Metrics}
We use the evaluation protocol of~\cite{niemeyer2019occupancy}. We evaluate the reconstruction and completion using Intersection over Union (IoU) and Chamfer Distance (CD). For the tracking, we use the $l_2$ correspondence metric (Corr): given a reconstructed sequence and its ground truth, 
it computes the $l_2$ distance between the 3D trajectory of a point on the reconstructions and the trajectory of its corresponding point on the ground truth shapes; correspondence is extracted by a nearest neighbour search to the first frame. Similar to~\cite{niemeyer2019occupancy, tang2021learning, lei2022cadex, cao2024motion2vecsets}, every shape is normalised so the maximum edge length of it bounding box is $10$.

\subsection{Human Motion Sequences}
\begin{figure}[h]
    \centering
    \includegraphics[width=1.0\columnwidth]{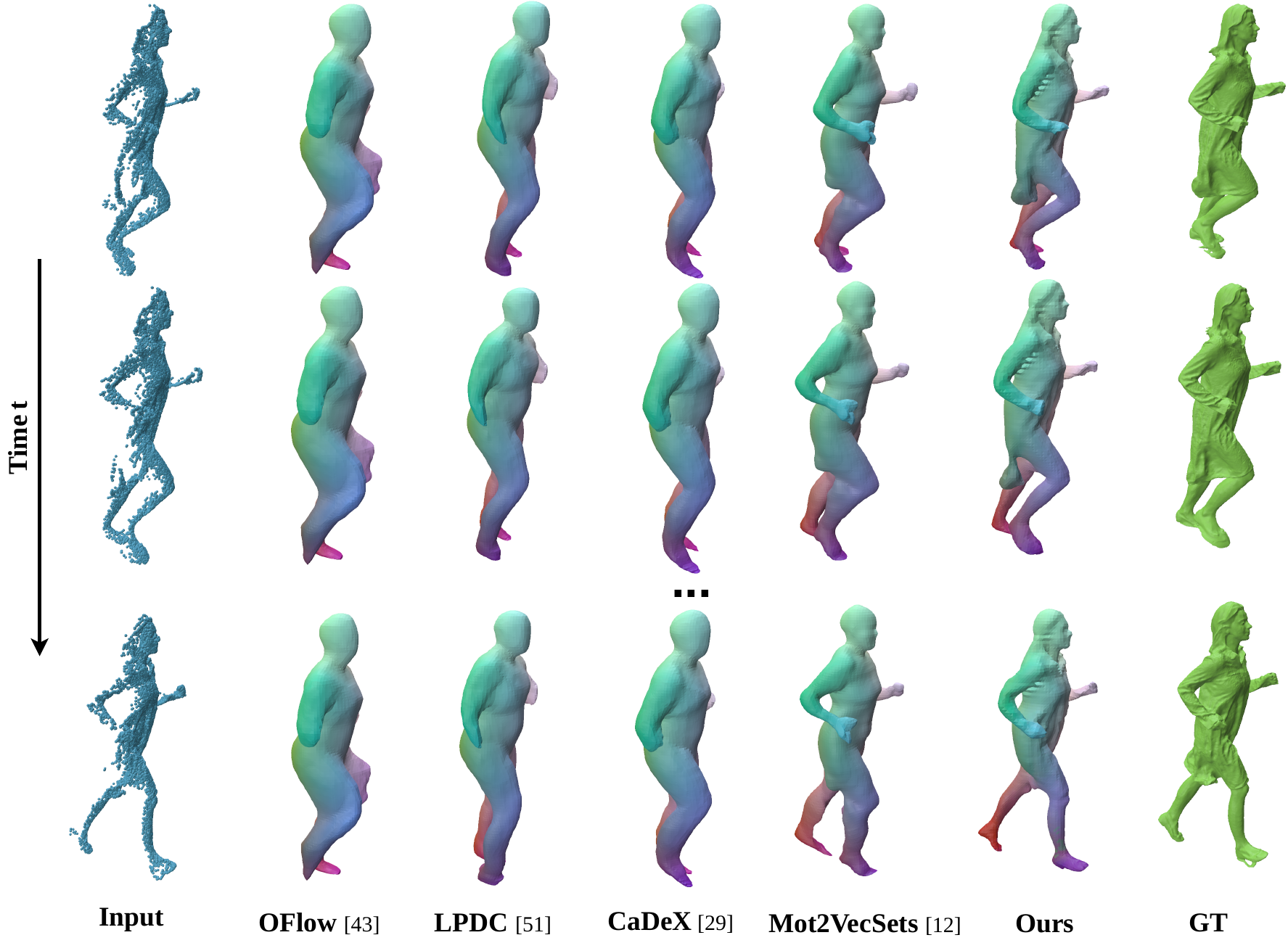}
    \caption{Qualitative comparison on Human motion reconstruction from monocular depth observations. Colors are defined on the first frame and transferred using predicted tracking. Ours is the only one that preserves observed geometric details.}
    \label{fig:qual_comp_h4do}
\end{figure}
\label{sec:human_rec}
Similar to~\cite{niemeyer2019occupancy, tang2021learning, lei2022cadex, cao2024motion2vecsets} the evaluation is conducted on sub-sequences of 17 frames. In practice, our method processes $5$ frames simultaneously. Therefore, we reconstruct $4$ sequences of $5$ frames with one frame overlap and extract the tracking using nearest neighbour search in 3D.

\noindent\textbf{D-FAUST} Tab.~\ref{tab:DFAUST_retrain} shows the quantitative results on the D-FAUST test set. It is split into motions and individuals unseen during training. In terms of reconstruction and completion quality, our method outperforms all competing methods on both folds. In terms of tracking quality, our method is on par with the other unsupervised method CaDeX, while being close to the best method Motion2VecSets. This shows that departing from the template+flow representation and estimating shape and motion in a coupled manner between frames allows to preserve more geometric details while keeping a competitive tracking quality.
\input{./tables/comparison_DFAUST_RETRAINED}
\input{./tables/comparison_cape}

\noindent\textbf{4DHumanOutfit and CAPE} Tab.~\ref{tab:H4DO} shows quantitative results on a subset of 4DHumanOutfit. Since the ground truth meshes are not registered to a common template (contrary to D-FAUST and CAPE), we use SMPL~\cite{loper2023smpl} fittings to evaluate the tracking. Our method outperforms both supervised and unsupervised methods in reconstruction and tracking, demonstrating superior cross-dataset generalisation. Fig.~\ref{fig:qual_comp_h4do} shows an example. 
OFlow, LPDC and CaDeX fail on this example. Motion2VecSets fails to capture observed details and the unconstrained flow representation causes distortions on unobserved surface parts. Conversely, our representation allows for both preservation of observed geometric detail and for minimising distortions of unobserved surface parts. 
We report similar results for the CAPE subset in Tab.~\ref{tab:CAPE}.

\input{./tables/comparison_4DHO}

\begin{figure}[h]
    \centering
    \includegraphics[width=1.0\columnwidth]{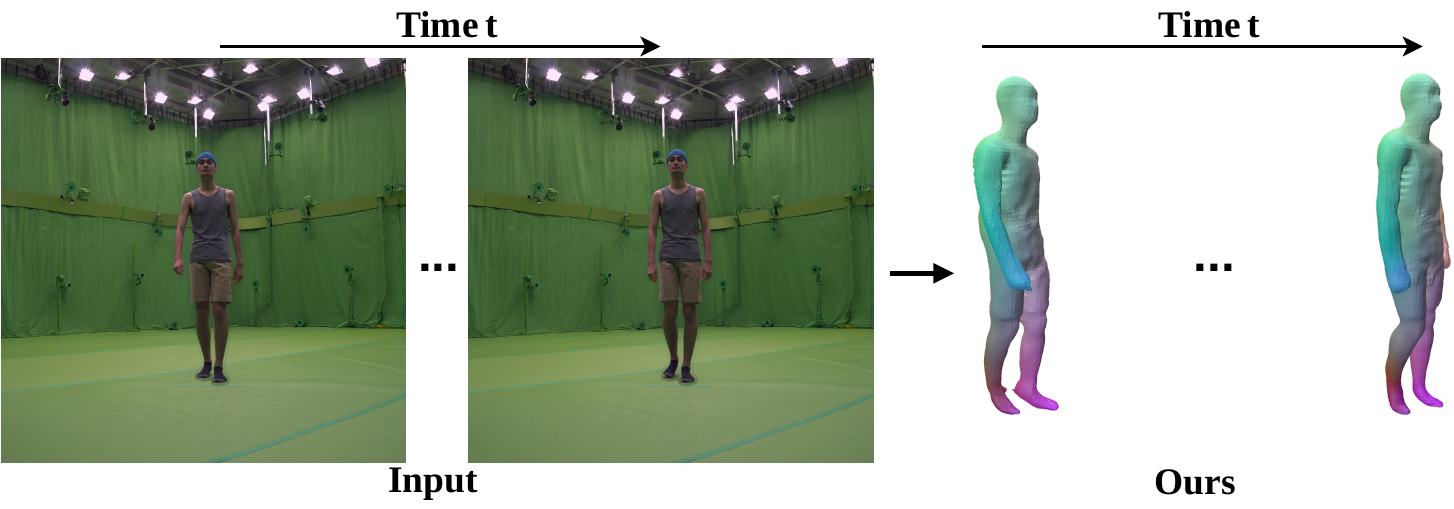}
    \caption{Results with depth map estimates from real monocular RGB images. First and tenth frame reconstructions are shown.}
    \label{fig:rebut_sapiens}
\end{figure}

Further qualitative experiments use depth estimates from RGB images using Sapiens~\cite{khirodkar2024sapiens} as input (see Fig.~\ref{fig:rebut_sapiens}). This is a challenging practical scenario, and our approach produces visually pleasing results.

\subsection{Animal Motion Sequences}
\label{sec:animal_rec}
  Tab.~\ref{tab:DT4DA_retrain} gives the quantitative results on the DT4D-A test set which is split into motions and individuals unseen during training. Our method outperforms all competing methods on completion
  showing that the near-isometric deformation assumption allows to generalise to different shape classes. 
\input{./tables/comparison_DT4D_RETRAINED}

\subsection{Ablation Studies}
\label{sec:ablation}
\begin{figure}[h]
    \centering
    \includegraphics[width=1.0\columnwidth]{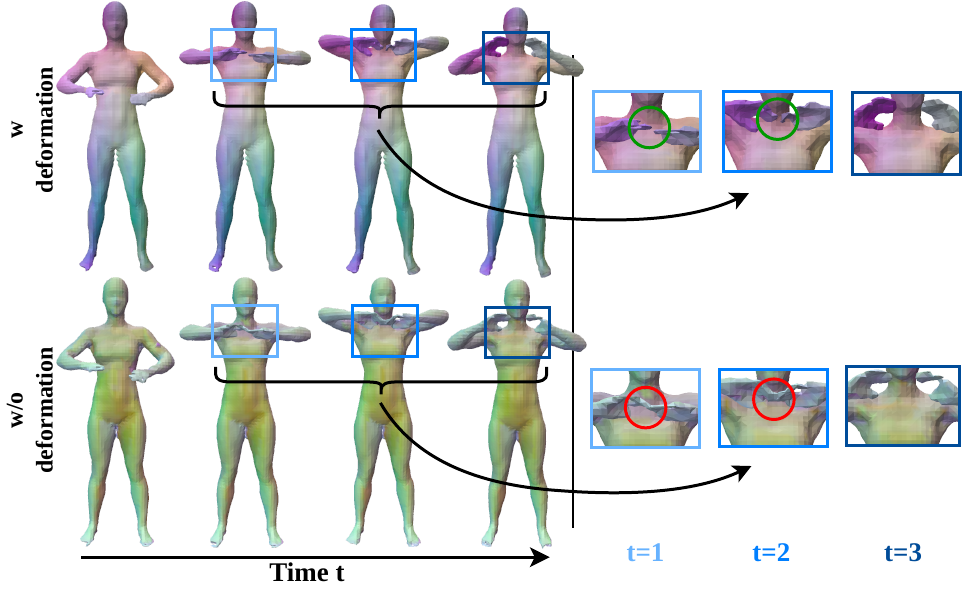}
    \caption{Geometry aligned features $(\mathcal{F}_i)_{i \in \{1,...,N\}}$ interpolated on the reconstructed surface and reduced using t-SNE~\cite{van2008visualizing} to $3$ channels and visualised as colors. 
    }
    \label{fig:tsne_features}
\end{figure}

\input{./tables/ablation_only_fusion}

\noindent\textbf{Benefit of the fusion mechanism} Tab.~\ref{tab:ablation_dfaust} shows that linking observations in feature space allows to improve the temporal coherence of the reconstructions. 
\noindent\textbf{Benefit of the deformation model} In addition to providing critical motion information, the deformation constraint improves the reconstructions' consistency as on the hands in Fig~\ref{fig:tsne_features}. Fig~\ref{fig:tsne_features} (left) shows the learnt geometry aligned features $(\mathcal{F}_i)_{i \in \{1,...,N\}}$ as colors
. When trained with the deformation constraint, these features encompass a temporal dimension: corresponding shape parts have the same color across time. Quantitative results are in Sec.~\ref{sec:ablation_deformation}.

%% file: tables/comparison_DFAUST_RETRAINED.tex
\begin{table}[ht]
    \resizebox{1.0\columnwidth}{!}{
    \begin{tabular}{@{}ccccccc}
    \toprule
    \multirow{1}{*}{Fold} & \multirow{1}{*}{Unsup.} &  \multirow{1}{*}{Method} & IoU $\uparrow$ & CD $\downarrow$ & Corr $\downarrow$\\  \midrule
    & \xmark  & \multicolumn{1}{c|}{LPDC~\cite{tang2021learning}} & {76.04}\% & {0.0928} & {0.1176}\\
    
    & \xmark& \multicolumn{1}{c|}{Mot2VecSets~\cite{cao2024motion2vecsets}} & {87.87}\% & {0.0410} & \textbf{0.1014}\\

    \multirow{3}{*}{\thead{Unseen \\ Motion}} & \cmark  & \multicolumn{1}{c|}{OFlow~\cite{niemeyer2019occupancy}} & {75.20}\% & {0.0993} & {0.1648}\\
    & \cmark& \multicolumn{1}{c|}{CaDeX~\cite{lei2022cadex}}  & {80.80}\% & {0.0738} &  \underline{0.1191}\\
    
    &\cmark &\multicolumn{1}{c|}{Ours} &  \underline{\textbf{90.78\%}} & \underline{\textbf{0.0323}} & {0.1342} \\ 
    
    \hline
    
    & \xmark  & \multicolumn{1}{c|}{LPDC~\cite{tang2021learning}} & {69.13}\% & {0.1024} & {0.1392}\\
    
    & \xmark& \multicolumn{1}{c|}{Mot2VecSets~\cite{cao2024motion2vecsets}} & {81.19}\% & {0.0522} & \textbf{0.1155} \\

    \multirow{3}{*}{\thead{Unseen \\ Individual}} &\cmark  & \multicolumn{1}{c|}{OFlow~\cite{niemeyer2019occupancy}}  & {65.59}\% & {0.1193} & {0.2050}\\
    & \cmark& \multicolumn{1}{c|}{CaDeX~\cite{lei2022cadex}} & {72.08}\% & {0.0892} & {0.1480}\\
    
    & \cmark&  \multicolumn{1}{c|}{Ours} & \underline{\textbf{90.30}\%} & \underline{\textbf{0.0290}} & \underline{{0.1245}}\\

    \bottomrule
    \end{tabular}}
     \caption{
     Quantitative 4D shape reconstruction results from monocular depth on D-FAUST~\cite{bogo2017dynamic} (all methods are retrained). Best overall in bold, best amongst unsupervised methods underlined.}
    \label{tab:DFAUST_retrain}

\end{table}

%% file: tables/comparison_cape.tex
\begin{table}[ht]
    \resizebox{1.0\columnwidth}{!}{
    \begin{tabular}{@{}cccccc}
    \toprule
    {Unsup.}  &  {Method} & IoU $\uparrow$ & CD $\downarrow$ & Corr $\downarrow$ \\ 
    \midrule
    \xmark  & \multicolumn{1}{c|}{{LPDC}~\cite{tang2021learning}} & {55.51}\% & {0.2085} & {0.4221}  \\
    
    \xmark& \multicolumn{1}{c|}{{Motion2VecSets}~\cite{cao2024motion2vecsets}} & {71.94}\% & {0.1140} & {0.3617}  \\

    \cmark & \multicolumn{1}{c|}{{OFlow}~\cite{niemeyer2019occupancy}} & {50.42}\% & {0.2696} & {0.5351}  \\
    \cmark& \multicolumn{1}{c|}{{CaDeX}~\cite{lei2022cadex}}  & {57.47}\% & {0.2167} &  {0.4228}  \\
    
    \cmark& \multicolumn{1}{c|}{{Ours}} & \underline{\textbf{85.85\%}} & \underline{\textbf{0.0560}} & \underline{\textbf{0.3084}} \\

    \bottomrule
    \end{tabular}}
     \caption{Quantitative 4D shape reconstruction results from monocular depth on CAPE~\cite{ma2020learning} (all methods are retrained). Best overall in bold, best amongst unsupervised methods underlined.}
    \label{tab:CAPE}

\end{table}

%% file: tables/comparison_4DHO.tex
\begin{table}[ht]
    \resizebox{1.0\columnwidth}{!}{
    \begin{tabular}{@{}cccccc}
    \toprule
    {Unsup.} &  {Method} & IoU $\uparrow$ & CD $\downarrow$ & Corr $\downarrow$ \\ 
    \midrule
    \xmark  & \multicolumn{1}{c|}{LPDC~\cite{tang2021learning}} & {58.86}\% & {0.1897} & {0.2739}  \\

    \xmark& \multicolumn{1}{c|}{{Motion2VecSets}~\cite{cao2024motion2vecsets}} & {71.64}\% & {0.0997} & {0.2583} \\
    
    \cmark  & \multicolumn{1}{c|}{{OFlow}~\cite{niemeyer2019occupancy}} & {56.03}\% & {0.2154} & {0.3299}  \\
    \cmark& \multicolumn{1}{c|}{{CaDeX}~\cite{lei2022cadex}}  & {61.28}\% & {0.1804} &  {0.2837}  \\
    
    \cmark& \multicolumn{1}{c|}{{Ours}} & \underline{\textbf{83.14\%}} & \underline{\textbf{0.0584}} & \underline{\textbf{0.2410}} \\

    \bottomrule
    \end{tabular}}
     \caption{Quantitative 4D shape reconstruction results from monocular depth on 4DHumanOutfit~\cite{armando20234dhumanoutfit} (all methods are retrained). Best overall in bold, best amongst unsupervised methods underlined.}
    \label{tab:H4DO}

\end{table}

%% file: tables/comparison_DT4D_RETRAINED.tex
\begin{table}[ht]
    \resizebox{1.0\columnwidth}{!}{
    \begin{tabular}{@{}cccccc}
    \toprule
    \multirow{1}{*}{Fold} & \multirow{1}{*}{Unsup.} &  \multirow{1}{*}{Method} & IoU $\uparrow$ & CD $\downarrow$ & Corr $\downarrow$\\  \midrule
    & \xmark  & \multicolumn{1}{c|}{LPDC~\cite{tang2021learning}} & {53.24}\% & {0.3961} & {0.4452}\\
    
    & \xmark& \multicolumn{1}{c|}{Mot2VecSets~\cite{cao2024motion2vecsets}} & {73.84}\% & {0.1790} & {0.4221}\\

    \multirow{3}{*}{\thead{Unseen \\ Motion}} & \cmark & \multicolumn{1}{c|}{OFlow~\cite{niemeyer2019occupancy}} & {67.29}\% & {0.2643} & {0.3812}\\
    & \cmark& \multicolumn{1}{c|}{CaDeX~\cite{lei2022cadex}}  & {76.57}\% & {0.1735} & \underline{\textbf{0.2970}}\\
    
    &\cmark& \multicolumn{1}{c|}{Ours} &   \underline{\textbf{76.73\%}} & \underline{\textbf{0.0990}} & {0.3564} \\ 
    
    \hline
    
    & \xmark & \multicolumn{1}{c|}{LPDC~\cite{tang2021learning}} & {47.31}\% & {0.4710} & {0.4672} \\
    
    & \xmark& \multicolumn{1}{c|}{Mot2VecSets~\cite{cao2024motion2vecsets}} & \textbf{66.45}\% & {0.1971} & {0.4600} \\

    \multirow{3}{*}{\thead{Unseen \\ Individual}} &\cmark & \multicolumn{1}{c|}{OFlow~\cite{niemeyer2019occupancy}}  & {57.13}\% & {0.3994} & {0.4525} \\
    & \cmark& \multicolumn{1}{c|}{CaDeX~\cite{lei2022cadex}} &{64.87}\% & {0.2704} & \underline{\textbf{0.3558}}\\
    
    & \cmark& \multicolumn{1}{c|}{Ours} & \underline{66.32\%} & \underline{\textbf{0.1478}} & {0.4850}\\

    \bottomrule
    \end{tabular}}
    \caption{Quantitative 4D shape reconstruction results from monocular depth on DT4D-A~\cite{li20214dcomplete} (all methods are retrained). Best overall in bold, best amongst unsupervised methods underlined.}
    \label{tab:DT4DA_retrain}

\end{table}

%% file: tables/ablation_only_fusion.tex
\begin{table}[ht]
    \resizebox{1.0\columnwidth}{!}{
    \begin{tabular}{@{}cccc|ccc@{}}
    \toprule
     \multirow{2}{*}{\thead{Fusion \\ Mechanism}} & \multicolumn{3}{c|}{Unseen Motion} & \multicolumn{3}{c}{Unseen Individual} \\
    \cmidrule(l){2-7}
    
    & IoU $\uparrow$ & CD $\downarrow$ & Corr $\downarrow$ & IoU $\uparrow$ & CD $\downarrow$ & Corr $\downarrow$ \\  \midrule
    
    \xmark  & \textbf{91.73}\% & \textbf{0.0299} & {0.1382} & \textbf{90.35}\% & {0.0295} & {0.1293} \\
    \cmark & {90.78\%} & {0.0323} & \textbf{0.1342} & {90.30}\% & \textbf{0.0290} & \textbf{0.1245} \\
   
    \bottomrule
    \end{tabular}}
     \caption{Ablation of feature-fusion on D-FAUST~\cite{bogo2017dynamic}. Best in bold.}
    \label{tab:ablation_dfaust}

\end{table}

%% file: sec/conclusions.tex
We propose a novel representation of moving 3D shapes that combines neural distance fields with a near-isometric mesh deformation model. Our representation allows for high fidelity reconstructions and a precise tracking, as shown in experiments on 4D reconstruction from monocular depth videos. Our approach can generalize to different shape classes and displays impressive cross-dataset generalisation going beyond results reported in prior works.

When dealing with motions that significantly deviate from the ones seen during training, our approach can provide an inaccurate tracking. Since the tracking strategy is unsupervised, test-time optimisation can be used to deal with out-of-distribution motions.

%% file: sec/X_suppl.tex
\clearpage
\setcounter{page}{1}
\maketitlesupplementary

This supplementary material presents an additional comparison with Neural Parametric Models (NPMs)~\cite{palafox2021npms} in Sec.~\ref{sec:npm_comparison}, ablation results on the deformation constraint in Sec.~\ref{sec:ablation_deformation}, and implementation details in Sec.~\ref{sec:implementation_details}. Our code is available at : \href{https://gitlab.inria.fr/amerrouc/combining-neural-fields-and-deformation-models-for-non-rigid-3d-motion-reconstruction-from-partial-data}{https://gitlab.inria.fr/amerrouc/combining-neural-fields-and-deformation-models-for-non-rigid-3d-motion-reconstruction-from-partial-data}

\section{Comparison to NPMs}
\label{sec:npm_comparison}

\begin{figure}[h]
    \centering
    \includegraphics[width=1.0\columnwidth]{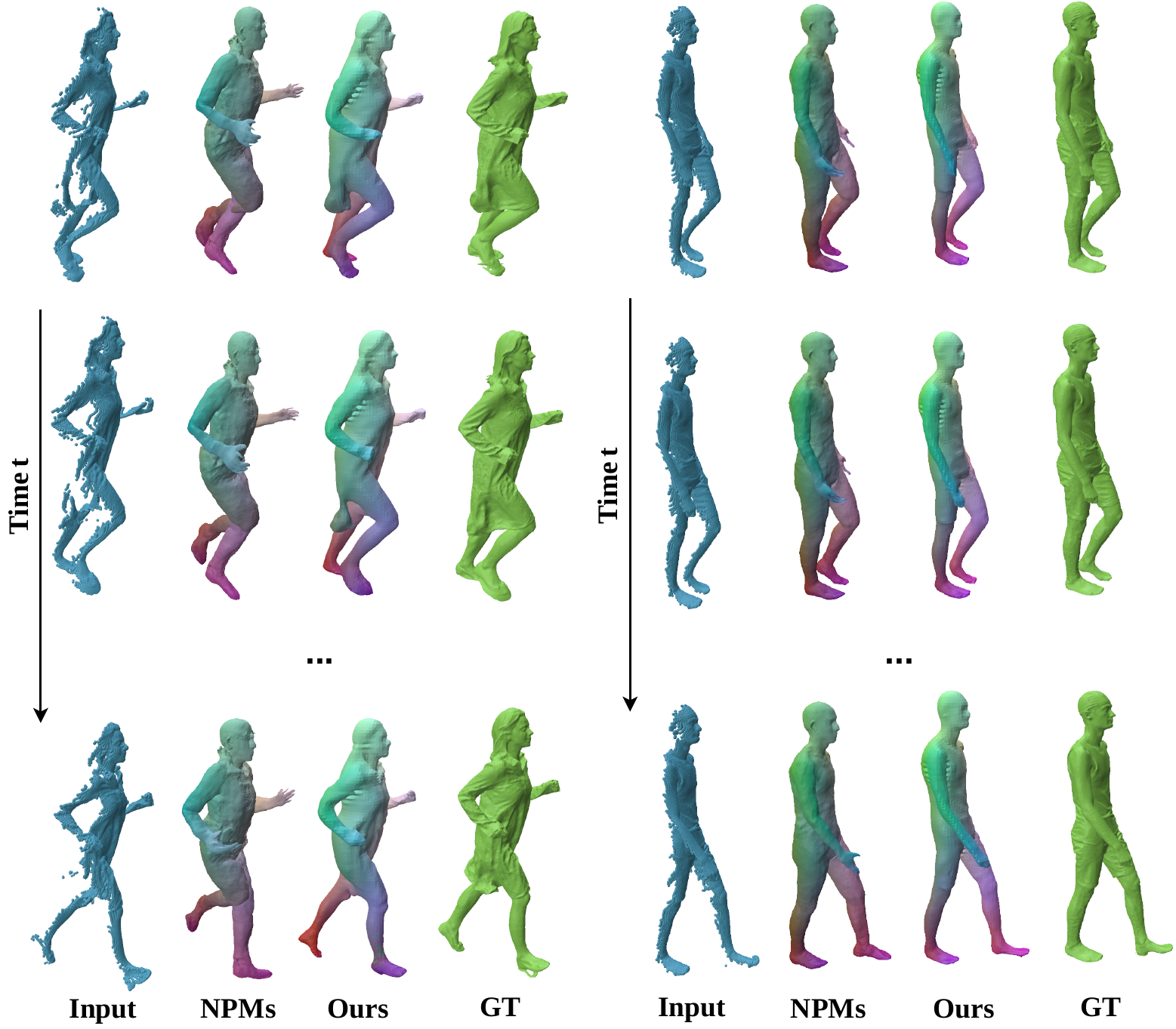}
    \caption{Qualitative comparison with NPMs~\cite{palafox2021npms} on Human motion reconstruction from monocular depth observations. Colors are defined on the first frame and transferred using predicted tracking. } 
    \label{fig:qual_comp_h4do_npm}
\end{figure}

We compare against model-based method Neural Parametric Models (NPMs)~\cite{palafox2021npms} on 3D motion reconstruction from monocular depth videos. NPMs learn disentangled pose and shape spaces from a dataset to which  partial observations of a moving shape can be fitted during inference through test-time optimisation. We use the pose and shape spaces pre-trained on human shapes from different datasets~\cite{li20214dcomplete, ma2020learning, mahmood2019amass} and provided by the authors. Our data-driven method trains on D-FAUST~\cite{bogo2017dynamic}. We tested on the 4DHumanOutfit~\cite{armando20234dhumanoutfit} dataset which consists of sequences of clothed human motions captured using a multi-camera platform. We use the multi-view mesh reconstructions as ground truth for the completion and SMPL~\cite{loper2023smpl} fittings as ground truth for the tracking. We synthetically generate monocular depth videos from the mesh sequences and use TSDF volumes as input for both our method and NPMs. Tab.~\ref{tab:H4DO_npm} presents the quantitative comparative results. Our method outperforms NPMs in both completion and tracking. Fig.~\ref{fig:qual_comp_h4do_npm} shows a qualitative comparison on two examples. Colors are defined on the first frame and transferred using predicted tracking. Our method achieves higher fidelity completions in both cases. Further, NPMs fail to infer the correct pose in the presence of loose clothing: the legs are crossed in the first example. This shows that model-based strategies struggle with examples that deviate from the shape-pose space hypothesis they consider. 

\input{./tables/comparison_4DHO_NPM}
\section{Ablation of the Deformation Constraint}
\label{sec:ablation_deformation}
We assess the benefit of the two main components of our method. First, the fusion mechanism (Sec.~\ref{sec:fusion}) where we ablate the feature-fusion in latent space. Second, the inter-frame deformation constraint (Sec.~\ref{sec:inter_frame_def_est}) where the network is restricted to the  feature-fusion based completion module. The added benefit of the fusion mechanism is shown in Tab.~\ref{tab:ablation_dfaust}. Tab.~\ref{tab:ablation_suppl} shows quantitative results of the ablation of the deformation constraint. The full model is, overall, on par in terms of reconstruction and completion quality while being augmented with crucial motion information. The deformation constraint also promotes the consistency of the reconstructions as shown in Fig.~\ref{fig:tsne_features}.

\input{./tables/ablation_supplementary}

\section{Implementation Details}
\label{sec:implementation_details}

\subsection{Feature-Fusion Based Completion}

\begin{figure}[h!]
    \centering
    \includegraphics[width=1.0\columnwidth]{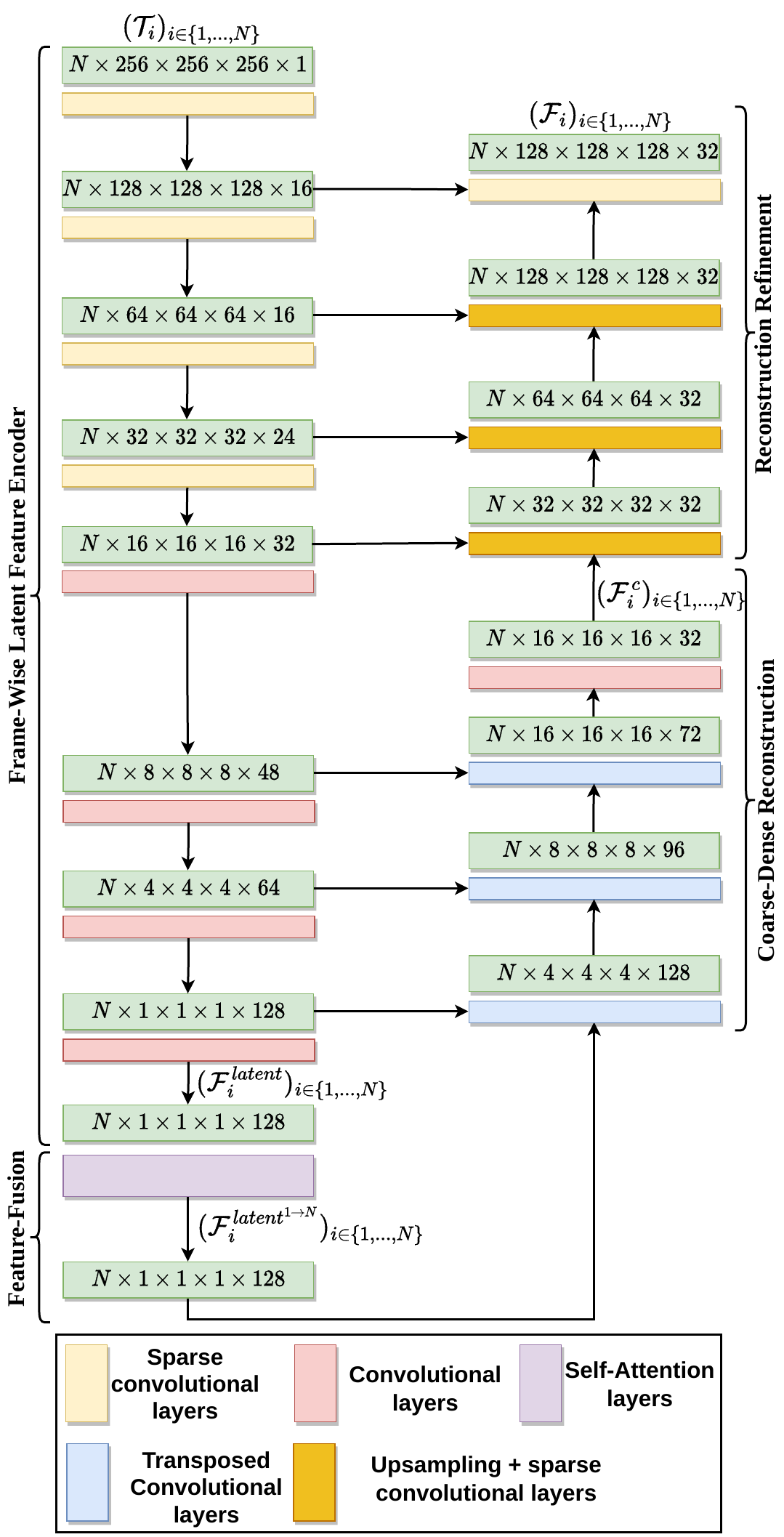}
    \caption{Architecture details of the feature extractor $F_{\Psi}$.}
    \label{fig:feature_extractor}
\end{figure}

\subsubsection{Architecture Details}

\paragraph{Feature Extractor $F_{\Psi}$} Fig.~\ref{fig:feature_extractor} gives more details about the feature extractor's architecture. For the feature-fusion, we employ self-attention with sinusoidal positional encoding. We use $2$ self-attention layers with $4$ attention heads.

\begin{figure}[h!]
    \centering
    \includegraphics[width=1.0\columnwidth]{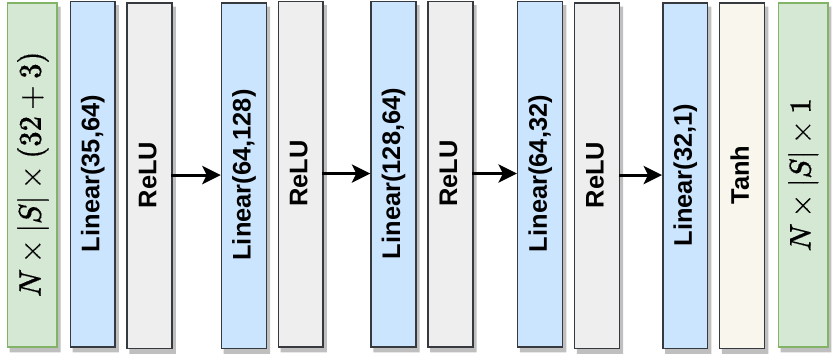}
    \caption{Architecture details of the MLP $S_{\Sigma}$.}
    \label{fig:MLP}
\end{figure}

\paragraph{MLP $S_{\Sigma}$} Fig.~\ref{fig:MLP} details $S_{\Sigma}$'s architecture. 

\subsubsection{Sampling Strategy for SDF Losses}
We use ground truth SDF samples generated from the mesh sequences in Eq.~\ref{sdf_coarse_loss} and Eq.~\ref{eik_coarse_loss} to supervise the completion task. For each frame $i$, we sample a total of $|S|$ points: $30\%$ are sampled uniformly in the bounding box $B$, while $70\%$ are sampled within a distance of $0.05$ to the mesh surface. In our experiments we fix $|S| = 50k$. The bounding box's extents are fixed to $(-0.5,-0.5,-0.5)$ and $(0.5,0.5,0.5)$.

\subsection{Inter-Frame Deformation Estimation}

\begin{figure}[h]
    \centering
    \includegraphics[width=0.9\columnwidth]{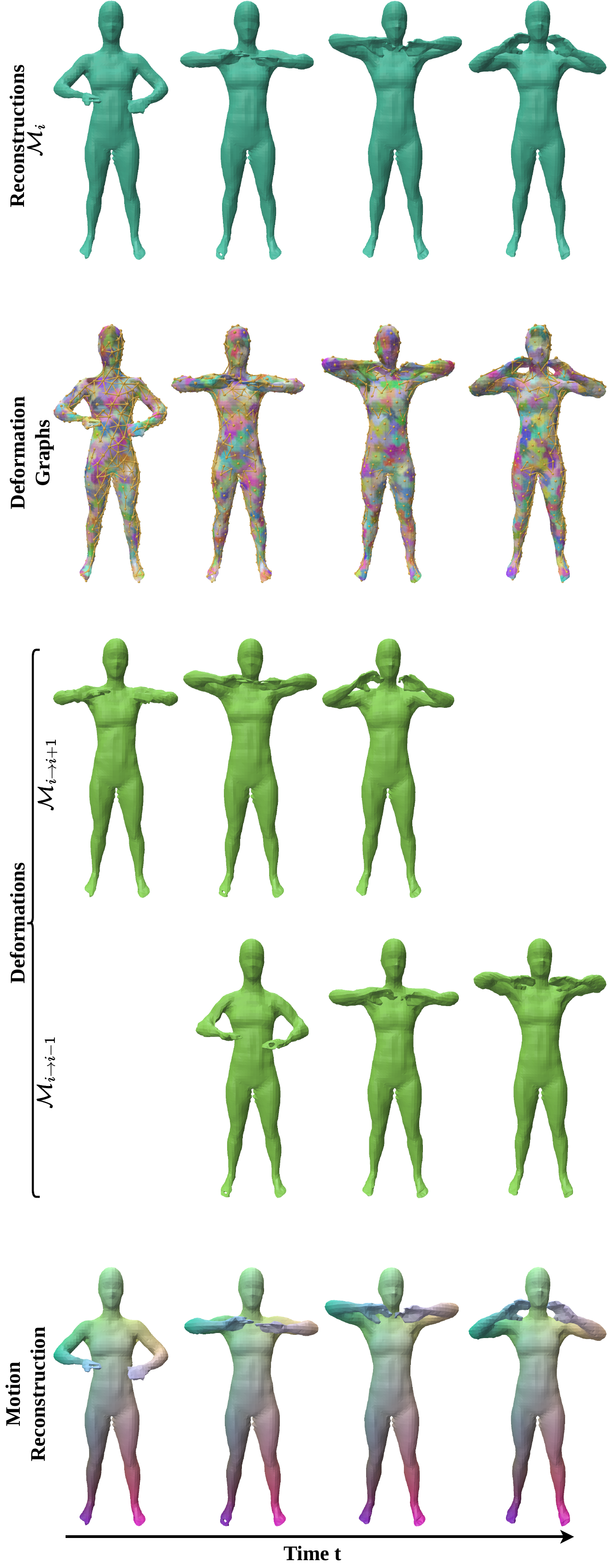}
    \caption{Our deformation guided tracking strategy. We fit the surfaces underlying our neural fields (top row), to a patch-wise near rigid deformation model; the second row shows the corresponding deformation graphs. A deformation decoder predicts these inter-frame deformations (third and fourth row). Given the deformations, we can extract a tracking using nearest neighbour search (bottom row).}
    \label{fig:def_unrowl}
\end{figure}

\subsubsection{Association Losses}
\label{sec:assos_losses_supp}

As explained in the main paper, we leverage two criteria on the association matrices. First, a cycle consistency criterion enforcing length $2$ and length $3$ cycle consistency for each sequence. It is implemented as follows :

{
\footnotesize
\begin{equation}
\label{cycle_loss_2}
\begin{split}
    l^{cycle}_2 &= \frac{1}{N-1}( \sum\limits_{i=1}^{N-1}{\|\Pi_{i \rightarrow i+1} (\Pi_{i+1 \rightarrow i} C^{i}) - C^{i}  \|^2_2 }
    \\&+  \sum\limits_{i=2}^{N}{\|\Pi_{i \rightarrow i-1} (\Pi_{i-1 \rightarrow i} C^{i}) - C^{i}  \|^2_2 }),
\end{split}
\end{equation}
}

{\footnotesize
\begin{equation}
\label{cycle_loss_3}
\begin{split}
    l^{cycle}_3 &= \frac{1}{N-2}(\sum\limits_{i=1}^{N-2}{\|\Pi_{i \rightarrow i+2} (\Pi_{i+2 \rightarrow i} C^{i}) - C^{i}  \|^2_2 }
    \\&+  \sum\limits_{i=3}^{N}{\|\Pi_{i \rightarrow i-2} (\Pi_{i-2 \rightarrow i} C^{i}) - C^{i}  \|^2_2 }),
\end{split}
\end{equation}}

{
\begin{equation}
\label{cycle_loss}
    l^{cycle} = l^{cycle}_2 + l^{cycle}_3,
\end{equation}}
where $\Pi_{i \rightarrow i+2} := \Pi_{i \rightarrow i+1}  \Pi_{i+1 \rightarrow i+2}$.

Second, a self-reconstruction criterion that identifies each patch in feature space to avoid many-to-one patch associations. It is implemented as follows:
\begin{equation}
\label{self_rec_loss}
 l^{rec} =  \frac{1}{N}(\sum\limits_{i=1}^{N}\|\Pi_{i \rightarrow i} C^{i}   - C^{i} \|^2_2),
\end{equation}
where $\Pi_{i \rightarrow i}$ is a self association matrix computed similarly to Eq.~\ref{pi_x_y} and Eq.~\ref{pi_y_x}. We also compute associations on the coarse geometry aligned features $(\mathcal{F}^{c}_i)_{i \in \{1,...,N\}}$ and optimise for these losses that we denote $l^{cycle}_{coarse}$ and $l^{rec}_{coarse}$. We promote these properties for coarse level features; in turn they will be inherited by finer level features~\ie by $(\mathcal{F}_i)_{i \in \{1,...,N\}}$. The loss $l^{associate}$ in Eq.~\ref{l_associoate}, integrating the coarse level association losses is implemented as follows:

\begin{equation}
\label{l_associoate_with_coarse}
    l^{associate} =   \lambda_3(l^{cycle} + l^{cycle}_{coarse}) + \lambda_4(l^{rec} + l^{rec}_{coarse}),
\end{equation}
where $\lambda_3, \lambda_4 \in \mathbb{R}$ are weights for loss terms.

\subsubsection{Deformation Model}
\label{sec:def_model_supp}

We use a patch-based mesh deformation model~\cite{cagniart2010free} to model inter-frame evolution. It decomposes a non-rigid deformation into patch-wise rigid deformations blended at the vertex level. Each mesh $\mathcal{M}$ is decomposed into non-overlapping surface patches $(P_k)_{1 \leq k \leq L}$ with their centers $\mathcal{C} = (c_k \in \mathbb{R}^3)_{1 \leq k \leq L}$ where $L$ is the number of patches. Each patch $P_k$ defines its rigid deformation \ie~ a rotation $R_k \in \mathbb{R}^{3 \times 3}$ and a translation $u_k \in \mathbb{R}^3$ as well as a blending function $\alpha_k(v)$ that depends on the euclidean distance of $v$ to $c_k$. The deformation model optimises for a rigidity constraint that implements a near isometric-deformation assumption by promoting consistent deformations between every patch $P_k$ and its neighbours $\mathcal{N}(P_k)$.  The rigidity constraint we use in Eq.~\ref{rig_loss} is implemented as follows:
{
\begin{equation}
    l_{rig}(\mathcal{M}) = \sum_{(P_k)_{1 \leq k \leq L}}{\sum_{P_j \in \mathcal{N}(P_k)}{\sum_{v \in P_k \bigcup P_j}{E^{kj}_v}}} \mbox{ with,}
\end{equation}}
\begin{equation}
        E^{kj}_{v\in P_k \bigcup P_j} =  (\alpha_k(v) + \alpha_j(v)) \| x_k(v) - x_j(v) \|^2_2,
\end{equation}
where $x_k(v)$ is the deformation defined by $P_k$ applied on $v$ \ie~$R_k (v - c_k) + u_k$. In our experiments we fixed the number of patches to $L=400$. Fig.~\ref{fig:def_unrowl} shows examples: the top row shows meshes and the second row shows their patch decomposition along with patch centers and patch adjacencies represented as a graph.

\subsubsection{Deformation Decoder}
To learn this deformation, we employ the deformation decoder introduced for static complete 3D shapes in~\cite{Merrouche_2023_BMVC}. It consists of a hierarchical graph convolutional network acting on the patch neighborhoods followed by an MLP. We use three patch levels in the hierarchical graph convolutional network: $20$, $50$ and $400$. It outputs the deformation model's parameters \ie~$(R_k \in \mathbb{R}^6)_{1 \leq k \leq L}$ and new center positions $(u_k\in \mathbb{R}^3)_{1 \leq k \leq L}$ for every patch of $\mathcal{M}_i$. To output the deformation parameters induced by the association matrix $\Pi_{i \rightarrow j}$~\ie the one that deforms $\mathcal{M}_i$ into $\mathcal{M}_j$, it takes as input patch centers $C^{i} \in \mathbb{R}^{L \times 3}$, patch-wise features $\mathcal{F}^{patch}_i \in \mathbb{R}^{L \times C}$, 
target centers $\Pi_{i \rightarrow j} \mathcal{C}^j \in \mathbb{R}^{L \times 3}$, and $\Pi_{i \rightarrow j} \mathcal{F}^{patch}_j\in \mathbb{R}^{L \times C}$. Applying the output deformations leads to the deformed shape $\mathcal{M}_{i\rightarrow j}$. The $6D$ representation of rotations~\cite{6D_rotation} is used.

Fig.~\ref{fig:def_unrowl} shows an example of reconstructions and inter-frame deformations computed by the deformation decoder in the case of a fast motion. The top row shows the reconstructions $\mathcal{M}_i$, the second row shows the deformation graphs of the patch-wise deformation model, and third and forth rows show the deformed shapes $\mathcal{M}_{i\rightarrow i+1}$ and $\mathcal{M}_{i\rightarrow i-1}$ respectively. Using the inter-frame deformations $\mathcal{M}_{i\rightarrow i+1}$, we extract the tracking using a nearest neighbour search as shown in the bottom row. Note that we make no hypothesis on the speed of the motion. 

\subsection{Training Details}
To allow for more stable learning, the network optimises for $l^{network}$ in gradual steps. First, the network only optimises for $l^{coarse}$ to obtain coarse reconstructions. After $N_1$ epochs, assuming that we have roughly located the coarse surface, the refinement step is activated and the network optimises for both $l^{coarse}$ and $l^{fine}$ \ie for $l^{fusion}$. After $N_2$ epochs, given that we have converged to good initial refined surfaces, the association search is activated; the network optimises for both $l^{fusion}$ and $l^{associate}$. Finally, after $N_3$ epochs, the deformation search \ie $l^{deform}$ is activated and the network optimises for $l^{network}$ until convergence. Tab.~\ref{tab:loss_weights_dfaust} and Tab.~\ref{tab:loss_weights_dt4d} detail this in terms of loss weights for the model trained on D-FAUST~\cite{bogo2017dynamic} and the one trained on DT4D-A~\cite{li20214dcomplete} respectively.

\input{./tables/loss_weights}
\input{./tables/loss_weights_dt4d}

We train the network with the Adam~\cite{kingma2014adam} optimizer and use gradient clipping. We use a learning rate of $10^{-3}$ during the first training epoch, $5 \times 10^{-4}$ between the $2^{nd}$ and the $60^{th}$ epoch, $2.5 \times 10^{-4}$ between the $60^{th}$ epoch and the $100^{th}$ epoch and $1.25 \times 10^{-4}$ after the $100^{th}$ epoch.

%% file: tables/comparison_4DHO_NPM.tex
\begin{table}[ht]
    \resizebox{1.0\columnwidth}{!}{
    \begin{tabular}{@{}cccccc}
    \toprule
    {Unsup.} &  {Method} & IoU $\uparrow$ & CD $\downarrow$ & Corr $\downarrow$ \\ 
    \midrule

    \xmark & \multicolumn{1}{c|}{{NPMs}~\cite{palafox2021npms}} & {69.84}\% & {0.1192} & { 0.2547} \\

    \cmark& \multicolumn{1}{c|}{{Ours}} & {\textbf{83.14\%}} & {\textbf{0.0584}} & {\textbf{0.2410}} \\

    \bottomrule
    \end{tabular}}
     \caption{Quantitative comparisons of 4D Shape Reconstruction from monocular depth sequences on the 4DHumanOutfit~\cite{armando20234dhumanoutfit} test set. Best in bold.}
    \label{tab:H4DO_npm}

\end{table}

%% file: tables/ablation_supplementary.tex
\begin{table}[ht]
    \resizebox{0.99\columnwidth}{!}{
    \begin{tabular}{@{}ccccc|ccc@{}}
    \toprule
     \multirow{2}{*}{\thead{Fusion \\ Mechanism}} & \multirow{2}{*}{\thead{Deformation \\ Constraint}} & \multicolumn{3}{c|}{Unseen Motion} & \multicolumn{3}{c}{Unseen Individual} \\
    \cmidrule(l){3-8}
    
    && IoU $\uparrow$ & CD $\downarrow$ & Corr $\downarrow$ & IoU $\uparrow$ & CD $\downarrow$ & Corr $\downarrow$ \\  \midrule
    
    \cmark & \xmark & \textbf{91.19}\% & \textbf{0.0309} & {-} & {89.48}\% & {0.0311} & {-} \\
    \cmark & \cmark & {90.78\%} & {0.0323} & \textbf{0.1342} & \textbf{90.30}\% & \textbf{0.0290} & \textbf{0.1245} \\
   
    \bottomrule
    \end{tabular}}
     \caption{Ablation result of the deformation constraint on D-FAUST~\cite{bogo2017dynamic}. ``-" means not applicable. Best in bold.}
    \label{tab:ablation_suppl}

\end{table}

%% file: tables/loss_weights.tex
\begin{table}[ht]

    \resizebox{1.0\columnwidth}{!}{
    \begin{tabular}{ |c||c|c|c|c|c|c| }
 \hline
 \thead{Train\\Epoch}  & \thead{$\leq 200(N_1)$} & \thead{$200<$ ,\\ $\leq 250$} & \thead{$250<$ ,\\ $\leq 400$} & \thead{$400(N_2)<$ ,\\ $\leq 430$}& \thead{$430<$ ,\\ $\leq 450$}  & \thead{$450(N_3)<$}\\
 \hline
 \hline
 $l^{SDF}_{coarse}$ & $2 \times 10^3$ & $2 \times 10^3$ & $2 \times 10^3$ & $2 \times 10^3$ & $2 \times 10^3$ & $2 \times 10^3$ \\
 \hline
 $l^{eikonal}_{coarse}$ & $4 \times 10$ & $4 \times 10$ & $4 \times 10$ & $4 \times 10$  & $4 \times 10$  & $4 \times 10$ \\
 \hline
 $l^{SDF}_{fine}$ & 0 & $2 \times 10^3$ & $2 \times 10^3$ & $2 \times 10^3$ & $2 \times 10^3$ & $2 \times 10^3$ \\
 \hline 
 $l^{eikonal}_{fine}$ & 0 & $4 \times 10^{-2}$ & $4 \times 10$ & $4 \times 10$ & $4 \times 10$ & $4 \times 10$ \\
 \hline
 \thead{$l^{cycle}$,\\$l^{cycle}_{coarse}$} & 0 & 0 & 0 & $10$ & $10^3$ & $10^3$ \\
 \hline
 \thead{$l^{rec}$,\\$l^{rec}_{coarse}$} & 0 & 0 & 0 & $10$ & $10^3$  & $10^3$\\
 \hline
 $l^{match}$ & 0 & 0 & 0 & 0 & 0 & $4 \times 10^3$\\
 \hline
 $l^{rigidity}$ & 0 & 0 & 0 & 0 & 0 & $4 \times 10^5$\\
 \hline
 $l^{surf}$ & 0 & 0 & 0 & 0 & 0 & $4 \times 10^2$\\
 \hline
  
\end{tabular}}
\caption{Loss weights for each loss term during training on the DFAUST~\cite{bogo2017dynamic} dataset.}
\label{tab:loss_weights_dfaust}
\end{table}

%% file: tables/loss_weights_dt4d.tex
\begin{table}[ht]

    \resizebox{1.0\columnwidth}{!}{
    \begin{tabular}{ |c||c|c|c|c|c|c| }
 \hline
 \thead{Train\\Epoch}  & \thead{$\leq 200(N_1)$} & \thead{$200<$ ,\\ $\leq 300$} & \thead{$300<$ ,\\ $\leq 400$} & \thead{$400(N_2)<$ ,\\ $\leq 450$}& \thead{$450<$ ,\\ $\leq 470$}  & \thead{$470(N_3)<$}\\
 \hline
 \hline
 $l^{SDF}_{coarse}$ & $2 \times 10^3$ & $2 \times 10^3$ & $2 \times 10^3$ & $2 \times 10^3$ & $2 \times 10^3$ & $2 \times 10^3$ \\
 \hline
 $l^{eikonal}_{coarse}$ & $4 \times 10$ & $4 \times 10$ & $4 \times 10$ & $4 \times 10$  & $4 \times 10$  & $4 \times 10$ \\
 \hline
 $l^{SDF}_{fine}$ & 0 & $2 \times 10^3$ & $2 \times 10^3$ & $2 \times 10^3$ & $2 \times 10^3$ & $2 \times 10^3$ \\
 \hline 
 $l^{eikonal}_{fine}$ & 0 & $4 \times 10^{-2}$ & $4 \times 10$ & $4 \times 10$ & $4 \times 10$ & $4 \times 10$ \\
 \hline
 \thead{$l^{cycle}$,\\$l^{cycle}_{coarse}$} & 0 & 0 & 0 & $10$ & $10^3$ & $10^3$ \\
 \hline
 \thead{$l^{rec}$,\\$l^{rec}_{coarse}$} & 0 & 0 & 0 & $10$ & $10^3$  & $10^3$\\
 \hline
 $l^{match}$ & 0 & 0 & 0 & 0 & 0 & $4 \times 10^3$\\
 \hline
 $l^{rigidity}$ & 0 & 0 & 0 & 0 & 0 & $4 \times 10^5$\\
 \hline
 $l^{surf}$ & 0 & 0 & 0 & 0 & 0 & $4 \times 10^2$\\
 \hline
  
\end{tabular}}
\caption{Loss weights for each loss term during training on the DT4D-A~\cite{li20214dcomplete} dataset.}
\label{tab:loss_weights_dt4d}
\end{table}